%% file: main.tex
\pdfoutput=1
\documentclass[acmsmall]{acmart}

\usepackage{graphicx} % Required for inserting images
\usepackage{color,xcolor}
\usepackage{url}
\usepackage{natbib}
\usepackage{comment}
\usepackage{mathrsfs}

\usepackage[normalem]{ulem}

\newtheorem{defn}{Definition}
\usepackage{amsmath,amsfonts}
\definecolor{zju_color}{RGB}{30,144,255}
\newcommand{\zd}[1]{{\bf\color{green}{ZD: #1}}}

\usepackage{array}
\newcolumntype{L}[1]{>{\raggedright\let\newline\\\arraybackslash\hspace{0pt}}m{#1}}
\newcolumntype{C}[1]{>{\centering\let\newline  \\\arraybackslash\hspace{0pt}}m{#1}}
\newcolumntype{R}[1]{>{\raggedleft\let\newline \\\arraybackslash\hspace{0pt}}m{#1}}

\AtBeginDocument{%
  }

\setcopyright{acmlicensed}
\copyrightyear{2018}
\acmYear{2018}
\acmDOI{XXXXXXX.XXXXXXX}

\acmJournal{JACM}
\acmVolume{37}
\acmNumber{4}
\acmArticle{111}
\acmMonth{8}

\begin{document}

\title{Knowledge Augmented Complex Problem Solving with Large Language Models: A Survey}

\author{Da Zheng}
\affiliation{
    \institution{Ant Group}
    \city{Beijing}
    \country{China}
}
\email{zhengda.zheng@antgroup.com}
\authornote{Equal contribution}
\authornote{Corresponding author}

\author{Lun Du}
\affiliation{
    \institution{Ant Group}
    \city{Beijing}
    \country{China}
}
\email{dulun.dl@antgroup.com}
\authornotemark[1]

\author{Junwei Su}
\affiliation{
    \institution{The University of Hong Kong}
    \city{Hong Kong}
    \country{China}
}
\email{junweisu@connect.hku.hk}

\author{Yuchen Tian}
\affiliation{
    \institution{Ant Group}
    \city{Beijing}
    \country{China}
}
\email{wanglian.tyc@antgroup.com}

\author{Yuqi Zhu}
\affiliation{
    \institution{Zhejiang University}
    \city{Hangzhou}
    \country{China}
}
\email{zhuyuqi@zju.edu.cn}

\author{Jintian Zhang}
\affiliation{
    \institution{Zhejiang University}
    \city{Hangzhou}
    \country{China}
}
\email{zhangjintian@zju.edu.cn}

\author{Lanning Wei}
\affiliation{
    \institution{Ant Group}
    \city{Beijing}
    \country{China}
}
\email{weilanning.wln@antgroup.com}

\author{Ningyu Zhang}
\affiliation{
    \institution{Zhejiang University}
    \city{Hangzhou}
    \country{China}
}
\email{zhangningyu@zju.edu.cn}
\authornotemark[2]

\author{Huajun Chen}
\affiliation{
    \institution{Zhejiang University}
    \city{Hangzhou}
    \country{China}
}
\email{huajunsir@zju.edu.cn}
\authornotemark[2]

\renewcommand{\shortauthors}{}


\begin{abstract}
Problem-solving has been a fundamental driver of human progress in numerous domains. 
With advancements in artificial intelligence, Large Language Models (LLMs) have emerged as powerful tools capable of tackling complex problems across diverse domains. 
Unlike traditional computational systems, LLMs combine raw computational power with an approximation of human reasoning, allowing them to generate solutions, make inferences, and even leverage external computational tools.
However, applying LLMs to real-world problem-solving presents significant challenges, including multi-step reasoning, domain knowledge integration, and result verification.
This survey explores the capabilities and limitations of LLMs in complex problem-solving, examining techniques including Chain-of-Thought (CoT) reasoning, knowledge augmentation, and various LLM-based and tool-based verification techniques.
Additionally, we highlight domain-specific challenges in various domains, such as software engineering, mathematical reasoning and proving, data analysis and modeling, and scientific research.
The paper further discusses the fundamental limitations of the current LLM solutions and the future directions of LLM-based complex problems solving from the perspective of multi-step reasoning, domain knowledge integration and result verification.

\end{abstract}



\maketitle

\section{Introduction} \label{sec:intro}
\input{intro}

\section{Definition of Complex Problem Solving} \label{sec:definition}
\input{definitions}

\section{Methodology} \label{sec:method}
\input{methods}

\section{Domains} \label{sec:domain}
\input{apps}

\section{Discussions and Future Directions} \label{sec:discuss}
\input{discussions}

\section{Related Works} \label{sec:related}
\input{related}

\section{Conclusions} \label{sec:concl}
\input{conclusion}

\bibliographystyle{ACM-Reference-Format}
\bibliography{refs}

\end{document}

%% file: intro.tex
% Motivation
The history of human civilization has been shaped by the ability of solving problems, ranging from constructing shelters in ancient times to unlocking the mysteries of the universe. 
For example, ancient astronomers calculated the Earth's size, while modern scientists predict weather using computational models. With technological advancements, humanity has gradually shifted from relying solely on individual or collective intellect to leveraging powerful tools like computers to address increasingly complex challenges. This transition marks a paradigm shift in problem solving, evolving from purely human-centered approaches to a synergy between human ingenuity and computational capability.

% background
Today, LLM-based AI systems represent a groundbreaking advancement \cite{zhao2023survey,min2024imitate,li2025system,zhang2025and}.  Unlike traditional computers, which excel at precise calculations, LLMs simulate aspects of human reasoning, such as generating creative solutions and making contextual inferences. This positions LLMs as tools that combine computational power with an approximation of human thought to solve complex problems that are challenging to humans.
Similar to human problem-solving, LLMs can directly solve problems and generate final results; LLMs can leverage computers to solve problems by writing and executing code to get results.

\begin{figure*}
\centering
\includegraphics[width=1\textwidth]{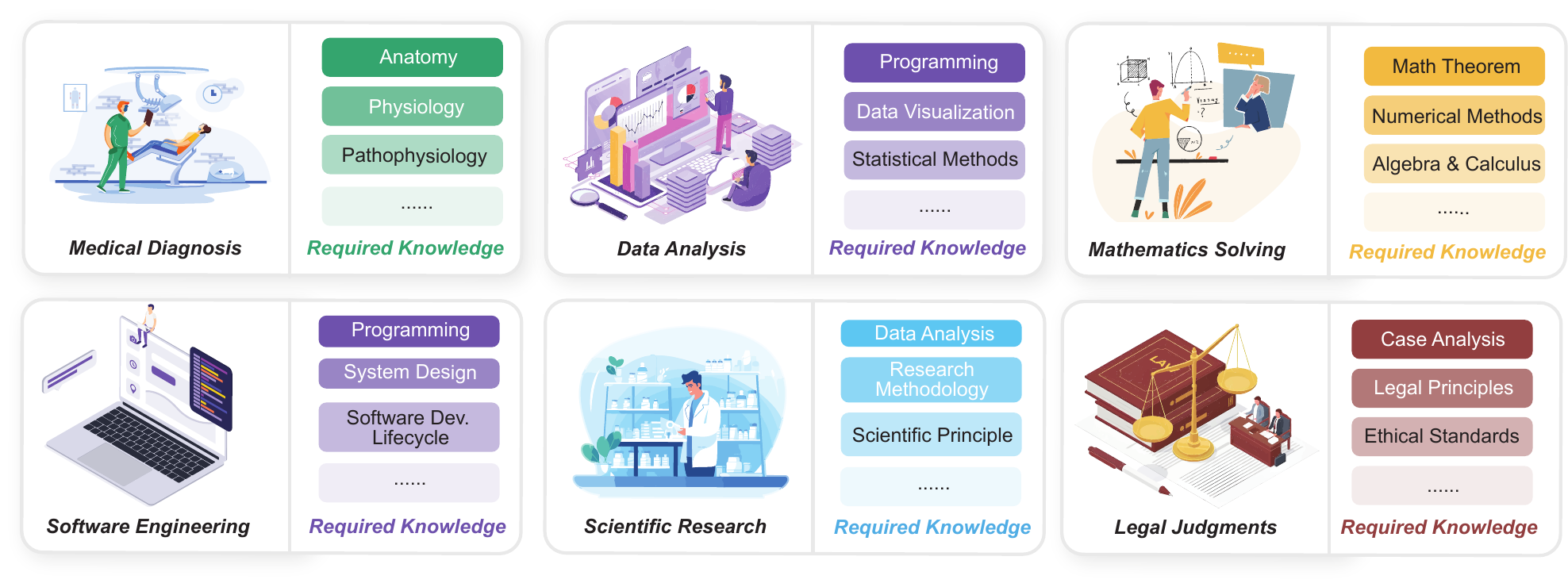}
\caption{Some scenarios for complex problem solving.}
\label{fig:domains}
\end{figure*}

The scope of complex problem solving spans a wide range of domains, encompassing challenges that touch virtually every aspect of human society (Figure \ref{fig:domains}). For instance, designing robust software system architectures requires balancing scalability, reliability, and user needs, while proving mathematical theorems demands rigorous logical reasoning and abstraction. In the realm of data science, building accurate models to interpret vast datasets is essential for informed decision-making. Similarly, drug discovery involves navigating intricate molecular interactions to identify effective therapies, and constructing physical models enables us to simulate and understand natural phenomena. These examples highlight the diversity of complex problems humanity strives to solve, each requiring a blend of domain expertise, reasoning, and creativity.

Solving these real-world complex problems involves leveraging domain knowledge or experience and progressing through multiple reasoning steps to arrive at a final solution. In the community, mathematical reasoning is frequently studied as a representative form of complex problem-solving and current research predominantly focuses on the mathematical reasoning problems with definitive final answers. In contrast, mathematical theorem proving tasks -- which are more representative of challenges encountered in higher education and research -- are often overlooked because they typically lack a single final answer to verify.
In practice, many real-world complex problems are even more challenging than mathematical reasoning tasks. First, these problems are inherently difficult to verify. For instance, in data science, numerous modeling techniques can be applied to the same dataset, yet their performance may vary greatly. Moreover, the effectiveness of a model is highly context-dependent, differing across problems and datasets. This variability makes it difficult to determine the optimal solution for any given modeling task.
Second, solving such real-world problems requires substantial domain expertise. Using data modeling again as an example, one must not only understand the nuances of the data but also be proficient in a wide range of modeling techniques to achieve strong performance.

Solving real-world complex problems requires three key components: \textit{multi-step reasoning}, \textit{domain knowledge}, and \textit{result verification}. This problem-solving process presents multiple challenges when LLMs are applied to real-world problems.
\begin{itemize}
    \item \textbf{Multi-step reasoning}: Solving a complex problem requires taking multiple steps to reach the final outcome. The size of the search space is largely determined by the number of steps needed to solve a complex problem, and can grow exponentially as the number of reasoning steps increases. This makes it challenging to identify the correct path to the final result. In addition, any errors that occur in the search process can propagate and lead to an incorrect result.
    \item \textbf{Domain knowledge}: Knowledge plays a crucial role in guiding the problem solver through the search space, helping to identify the next step or recognize when the solution has been reached.
    %Additionally, knowledge can create shortcuts in the reasoning process, enabling solutions to be found more efficiently by reducing the number of required steps and minimizing exploration complexity.
    Domain-specific applications, such as machine learning tasks and mathematical theorem proving tasks, typically require to utilize long-tail domain knowledge while it is well-known that LLMs cannot master low-tail knowledge well \cite{sun2024head}. This requires an LLM-based system to take extra care to master domain knowledge and reliably retrieve and apply the required knowledge to solve problems.
    \item \textbf{Result verification}: Each step must be carefully evaluated to assess whether it contributes to a correct solution or whether the entire solutions can solve the given problem. This evaluation can be particularly challenging in many applications where standard outcomes or predefined solution procedures are lacking. The difficulty is even greater for open-ended problems with ill-defined goals, such as those found in scientific research and data mining.
\end{itemize}

% CoT reasoning
Recent development of LLMs have demonstrated their strong reasoning capabilities on some complex problems that have
well-defined goals and whose results can be easily verified, making it ideal for tasks
like mathematical reasoning and competitive coding challenges. 
Chain-of-Thought (CoT) reasoning
is the major technique to solve multi-step reasoning \cite{wei2022chain, zhang2025igniting, yang2024buffer,chen2025towards}.
There is an inference scaling law in CoT reasoning that the likelihood of finding a correct solution improves
significantly as the number of CoT paths increases \cite{cobbe2021training} and it is often
possible to generate correct solutions for many challenging problems with a sufficient number of CoT paths \cite{brown2024large}.
Because the target applications, such as mathematical reasoning and competitive coding,
are easily verifiable, many works \cite{kumar2024training, deepseekai2025deepseekr1} are
using reinforcement learning to train LLMs to improve their reasoning capabilities for these applications \cite{li2025limr}.
The release of GPT-o1 by OpenAI \cite{openai2024gpto1} and DeepSeek-R1 \cite{deepseekai2025deepseekr1} showcases the potential of this CoT reasoning approach \cite{zeng2025revisiting}.

While CoT reasoning is an important technique to solve complex problems, it is necessary to adopt an agentic approach that enables access to external knowledge bases and the use of verification tools to further improve LLMs' capability in solving real-world complex problems. Previous studies have shown that LLMs have difficulty retaining long-tail knowledge \cite{sun2024head}, and domain-specific knowledge often falls into this category. It is essential to make external knowledge integration for knowledge-intensive tasks such as scientific discovery \cite{deepresearch}, mathematical theorem proving \cite{welleck2022natural}, and data science \cite{guo2024dsagent}, where domain expertise is critical for accurate and informed decision-making. Knowledge can be retrieved from documents through techniques like RAG \cite{gao2023retrieval} and GraphRAG \cite{edge2024local, han2024retrieval}, or by leveraging knowledge graphs constructed from document collections \cite{liang2024kag}. Additionally, agents may interact with humans to acquire domain knowledge directly \cite{do2024facilitating,zhang2025breaking}.
% Evaluation
Result verification is also essential to ensure valid solutions from large language models (LLMs), both during training and inference. Reasoning-focused LLMs
are often trained with synthetic data, which requires a verifier to select high-quality data
for model training \cite{cobbe2021training}. During inference, the inference scaling law
highlights the need for a verifier to identify the correct solution among multiple candidates
\cite{brown2024large}. Various types of verifiers can be employed for this purpose, including
LLM-as-a-judge approaches \cite{gu2025surveyllmasajudge}, symbolic reasoning tools ~\citep{agent_factool}, and even experimental validation systems \cite{arxiv24_autokaggle}.

% Practical applications
Despite the significant advancements in LLMs for complex problem solving, each domain presents its own unique challenges when applying LLMs to practical applications. Take some domains in
Figure \ref{fig:domains} as examples.
In software engineering, LLMs are tasked with generating or modifying code within large code repositories for bug fixings and new feature implementations. This requires them not only to reason about code generation but also to have a comprehensive understanding of the entire codebase and project requirements \cite{yang2024swe}. Furthermore, software development demands not just code correctness but also optimization in terms of computational efficiency and memory usage \cite{shypula2023learning}, adding an additional layer of complexity to the evaluation process.
Mathematics encompasses two primary types of tasks: calculations and provings. While extensive data are available for basic arithmetic and computational tasks, data scarcity remains a significant challenge in advanced mathematics, particularly in higher education and research \cite{glazer2024frontiermath}. To address this limitation, it is essential to leverage domain knowledge more effectively for data synthesis to mitigate the impact of data scarcity and utilize existing mathematical knowledge, such as theorems, to improve mathematical proving. In addition, mathematical theorem proving usually lack an effective way to verify the proving solution, making it difficult to train LLM models to generate rigorously correct mathematical reasoning solutions.
Data science involves working with large datasets, yet task descriptions often lack sufficient details about the distribution of input data, making it challenging for LLMs to generate the most suitable solutions to model the large datasets well \cite{arxiv24_mlebench}.
This also complicates the evaluation of LLM-generated outputs, necessitating multi-level assessments. Furthermore, leveraging a comprehensive knowledge base of data modeling techniques is crucial for developing more effective methods to tackle complex data science problems.
Scientific research often involves open-ended problems, which prevents us from training LLMs to solve scientific problems directly. One potential solution is to involve humans in the process (human-LLM collaboration), allowing for iterative collaboration between humans and LLMs to explore existing scientific literature and human knowledge \citep{scherbakov2024emergence,susnjak2024automating, beltagy2019scibert,jin2023pubmed,lo2019s2orc,luo2024large}, generate novel ideas \cite{si2024llms, wang2024scipip, wang2024scimon, baek2024researchagent} and automate entire research pipelines \cite{lu2024aiscientist}.
These challenges highlight the need for further research into complex problem-solving that go beyond the current reasoning LLMs.

% Overview

This paper provides a high-level overview of the current advancements in LLMs for solving complex problems and goes beyond the literature of reasoning LLMs. Our goal is to review the key techniques developed for LLMs and how these methods are being applied to address the challenges in different domains. The paper is structured into four sections to discuss the current LLM research:
\begin{itemize}
    \item \textbf{Definition of complex problem solving}: We begin by formally defining complex problem solving from the perspectives of cognitive science and computational theory (Section \ref{sec:definition}).
    \item \textbf{Methodologies}: We examine the key methodologies in LLM research for solving complex problems, including multi-step reasoning, knowledge augmentations and result verifications (Section \ref{sec:method}).
    \item \textbf{Domains}: We explore complex problem solving across four domains --software engineering, data science, mathematics, and scientific research --highlighting the unique challenges in each and the solutions developed to address them (Section \ref{sec:domain}).
    \item \textbf{Current limitations and future directions}: We discuss the limitations of current research and propose potential directions for future studies (Section \ref{sec:discuss}).
\end{itemize}

%% file: definitions.tex
\begin{figure}
\centering
\includegraphics[width=0.8\textwidth]{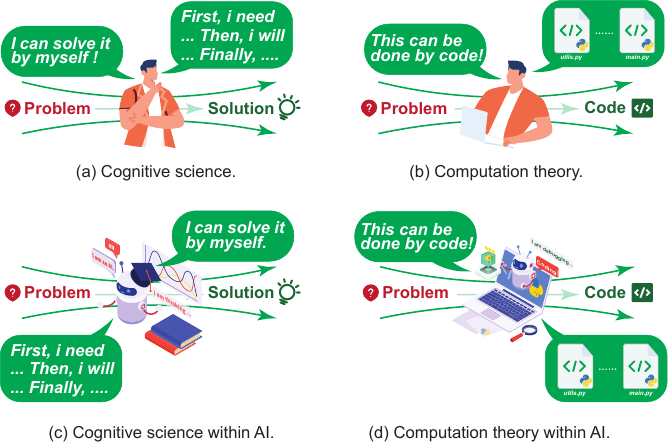}
\caption{Two paradigms of problem solving by humans and AI.}
\label{fig:cps}
\end{figure}

We can define complex problem solving from two perspectives: \textbf{cognitive science} and \textbf{computational theory}. Cognitive science investigates how humans use their inherent abilities to solve problems. Computational theory, by contrast, explores how to leverage machines for problem solving, emphasizing the design of algorithms to automate intricate computations.
When considering the role of LLMs in addressing complex problems, two potential paradigms emerge: (1) Direct problem solving: LLMs autonomously generate solutions akin to human experts. \footnote{+wln+: generate plans and manage the tools to solve the problems like human experts. (These components will be mentioned in the following methodologies and domains.)}; (2) Leveraging computational systems for complex problems: LLMs extract and define the computational components of a problem, utilizing traditional computers to execute intensive calculations while focusing on designing solutions and orchestrating processes.
With these paradigms in mind, this section will delve deeper into how CPS is defined in the frameworks of cognitive science and computational theory.

\subsection{Definitions}
%We first define \textbf{Problem} based on insights from cognitive science ~\citep{simon1971human,wang2010cognitive} and computational theory \citep{garey1979computers}.
\begin{defn}[\textbf{\emph{Problem}}]
  A problem $\Pi (\mathcal{X}, \mathcal{Y}, \mathcal{P})$ is described by (1) a description of its parameters $\mathcal{X}$, and (2) a statement (i.e., a predicate logic) $\mathcal{P}(Y; X)$ that characterizes the properties the solution must satisfy.
  Formally, the \textbf{goal} set  is defined as $\mathcal{G} = \{Y \in \mathcal{Y} \;|\; \mathcal{P}(Y; X)\}$, where $\mathcal{Y}$ is the space of final results, and $\mathcal{P}$ is a predicate logic that means $\mathcal{P}(Y; X)$ represents the property that a final result $Y$ should satisfy when $\mathcal{X} = X$.  An \textit{instance} $\pi$ of the problem is obtained by specifying particular values for all the problem parameters, i.e., $\pi := \Pi(\mathcal{X} = X)$. 
\end{defn} \label{def:1}

 A problem can be seen as a task that involves finding a solution from a set of possible candidates. The predicate $\mathcal{P}(Y; X)$ specifies the condition that an answer must satisfy to be considered valid. In different problems, the predicate $\mathcal{P}(Y; X)$ may either be well-defined or not. For instance, in the shortest path problem, the answer space $\mathcal{Y}$ consists of all possible paths, and the predicate $\mathcal{P}(Y; X)$ is well-defined, specifying that a final result $Y$ (a path) must satisfy the property of having the minimum total weight. In contrast, in data mining tasks, the goal is to discover insightful patterns within the data. However, what constitutes an ``insightful'' pattern is not clearly defined, making the predicate $\mathcal{P}(Y; X)$ more subjective and context-dependent.

Based on the definition of a problem, we can now formally define \textbf{problem solving} as the process of identifying a sequence of transformations that leads from an initial state to a goal state.

\begin{defn}[\textbf{\emph{Problem Solving}}]
\label{defn:problem-solving}
Problem solving is the process of finding a solution trace $T(\pi) \in \mathcal{T}_
{feasible} \subseteq \mathcal{T}$ for a problem instance $\pi$, where $\mathcal{T}_{feasible}$ is the set of all possible solution traces, formally defined as:
\[
\begin{split}
\mathcal{T}_{feasible} := \{&X \rightarrow O_1 \rightarrow \ldots \rightarrow O_\kappa \rightarrow Y | \\
&X \in \mathcal{X},\; Y \in \mathcal{G}, \kappa \in \mathbb{N}^+,\forall_{1 \le i \le \kappa}  O_i \in \mathcal{O}\},
\end{split}
\],
$\mathcal{T}$ is the set of all possible traces:
\[
\mathcal{T} := \{X \rightarrow O_1 \rightarrow \ldots \rightarrow O_\kappa \rightarrow Y | X \in \mathcal{X}, Y \in \mathcal{Y}, \kappa \in \mathbb{N}^+,  O_i \in \mathcal{O}\},
\],
and $\mathcal{O}$ is the set of all possible intermediate states during the problem-solving process.
\end{defn}

This definition emphasizes the iterative and state-dependent nature of problem solving, where intermediate states $O_i$ capture the evolving understanding or partial solutions leading to the final result $Y$.  However, the mechanisms driving state transitions and the constraints on feasible solution traces vary depending on the nature of the problem solver.

In a \textbf{human-centered} perspective, problem solving is inherently constrained by individual cognitive capabilities. The transition from one state to another is influenced not only by logical reasoning but also by domain knowledge, prior experience, and real-time feedback. As a result, different individuals may follow different paths within $\mathcal{T}_{feasible}$ based on their available cognitive resources. The formal definition is as follows:

\begin{defn}[\textbf{\emph{Human-Centered Problem Solving}}]
\label{defn:human-problem-solving}
Human-centered Problem solving is the process of finding a solution trace $T(\pi)$ for a problem instance $\pi$ by a person with cognitive capabilities $\mathcal{C}$, which include domain knowledge, logical reasoning, leveraging real-time feedback and other cognitive resources \cite{laird1996gentle}. The transition from an intermediate state $O_i$ to the next state $O_{i+1}$ is governed by a cognition-guided transition function:
\[
\Gamma: \mathcal{O} \times \mathcal{C} \to \mathcal{P}(\mathcal{O}),
\]
where $\mathcal{P}(\mathcal{O})$ is the power set of $\mathcal{O}$, representing all possible next states, and the transition function $\Gamma(O_i, C)$ determines the set of feasible next states given the solver’s cognitive capacity.
\end{defn}

Conversely, in a \textbf{computer-assisted} perspective, problem solving is approached from the lens of computational theory. Here, state transitions are governed by formal algorithms rather than cognitive capabilities. 

 %Since computational theory focuses on problem-solving processes assisted by computers, its definition differs from the human-centered perspective of cognitive science. Below, we provide definitions from both perspectives:

%In complex problem solving, a feasible solution trace $T(\pi) \in \mathcal{T}_{feasible}$ is often non-trivial and requires \textbf{multi-step reasoning} and \textbf{domain knowledge}. Unlike simple problems where a direct mapping from $X \to Y$ exists, complex problems typically involve a long sequence of intermediate states $O_1,..,O_k$, where each transformation $O_i \to O_{i+1}$ depends not only on reasoning ability but also on domain knowledge.

%\zd{should we add other properties described by the literature of cognitive science? for example, complex problems require knowledge and skill; they are dynamic and evolving, etc.}

%\begin{figure*}
% \includegraphics[width=1.0\textwidth]{images/CPS.png}
%\includegraphics[width=0.87\textwidth]{images/case-v1.pdf}
%\caption{Case Study: Example from the Field of Machine Learning}
%\label{fig:case}
%\end{figure*}

\begin{defn}[\textbf{\emph{Computer-assisted Problem Solving}}]
Problem solving is the process of designing \textit{algorithms} $\mathcal{A}$ to solve a problem $\Pi (\mathcal{X}, \mathcal{Y}, P)$. An algorithm is a finite sequence of instructions executable by a computer\footnote{Refers to a Turing machine or an equivalent computational architecture.} to produce solutions. Formally, an algorithm is defined as a quintuple:
\[
\mathcal{A} := (\mathcal{X}, \mathcal{Y}, \mathcal{O}, \delta, \sigma_0),
\]
where $\mathcal{X}$ is the input space, describing all possible parameters of the problem; $\mathcal{Y}$ is the output space, representing all potential solutions; $\mathcal{O}$ is the state space, containing all possible states during the execution of the algorithm; $\delta : \mathcal{X} \times \mathcal{O} \rightarrow \mathcal{O}$ is the state transition function, specifying how the algorithm transitions from one state to the next based on the input; $\sigma_0 \in \mathcal{O}$ is the initial state, representing the starting condition of the algorithm.
An \textit{algorithm} is said to solve a problem $\Pi$ if, for any instance $\pi$ of $\Pi$, it guarantees to produce a solution $Y \in \mathcal{G}$ that satisfies the predicate $P(Y;X)$.
\end{defn}
%\js{$\delta$ is commonly used to represent deviation. Shall we use another notation to represent transition?}
%This definition highlights the formal and deterministic nature of computer-assisted problem solving. 

By comparing the two definitions, we observe that they share a fundamental similarity: both focus on finding the steps to solve a problem. However, they differ in their emphasis. \textbf{Human-centered Problem Solving} mainly focuses on the process of solving a specific problem instance $\pi$, while \textbf{Computer-assisted Problem Solving} emphasizes designing relatively general algorithms to address a class of problems $\Pi$.

\subsection{Example}

\begin{figure}
\includegraphics[width=1\textwidth]{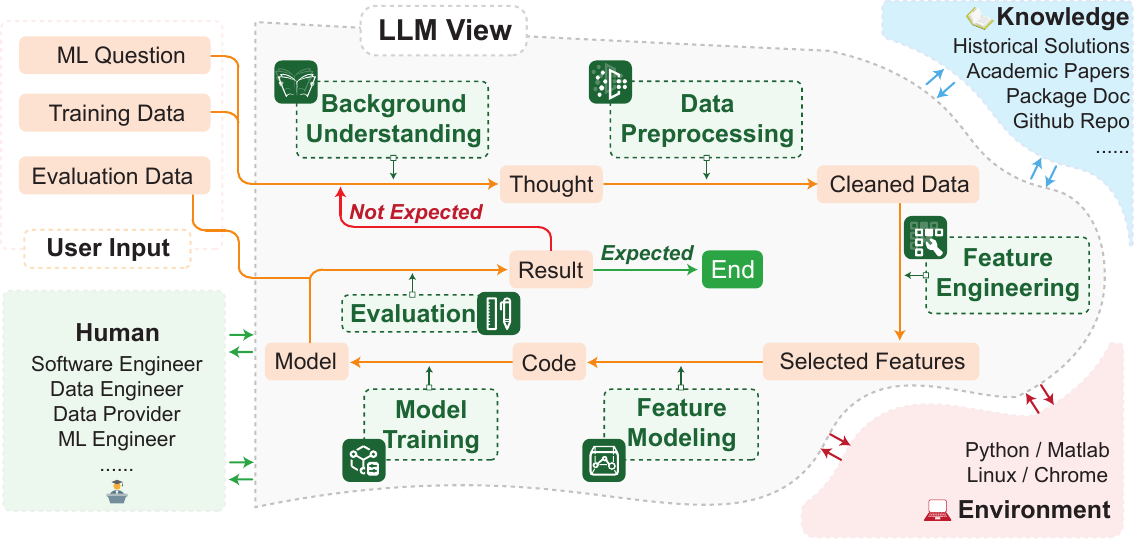}
\caption{Case Study: Example from the Field of Machine Learning}
\label{fig:case}
\end{figure}

Let's use machine learning tasks as an example (Figure \ref{fig:case}). Developing a high-quality machine learning model can be framed as a problem-solving process, where we seek a feasible solution trace $T(\pi) \in \mathcal{T}_{feasible}$.
Each step in this process corresponds to a state transition $O_i \to O_{i+1}$
 , driven by reasoning, domain knowledge, and iterative evaluation. Initially, we define the problem by identifying the task and structuring it into a machine learning formulation. Then, we transition through intermediate states by analyzing data, applying preprocessing techniques, and performing feature engineering. Once the data is processed, we select suitable modeling techniques and develop the model for training.
% Knowledge
To refine these transitions, domain knowledge plays a crucial role, guiding the selection of appropriate models and training strategies. Knowledge may originate from historical approaches, theoretical research, or expert intuition, shaping the feasible state space $\mathcal{T}_{feasible}$.
% Evaluation
Developing effective machine learning models requires multiple rounds of evaluation for each method, including both human assessments and experimental evaluations. Since machine learning models rely on learning data distributions from training data to make predictions, assessing the quality of a solution solely by examining it is challenging. Instead, empirical validation through human assessments and experimental testing determines the effectiveness of a model before convergence to an optimal solution $Y$.

%% file: methods.tex
%\js{I feel like that there is some deviation with respect to the granularity among the subsections. On one hand, agent and HLC can be considered distinct categories consisting of different techniques/frameworks? On the other hand,  COT can be considered a specific technique under the large family of prompt engineering such as tree of thought and graph of thought?}

\begin{figure}
\includegraphics[width=1.0\textwidth]{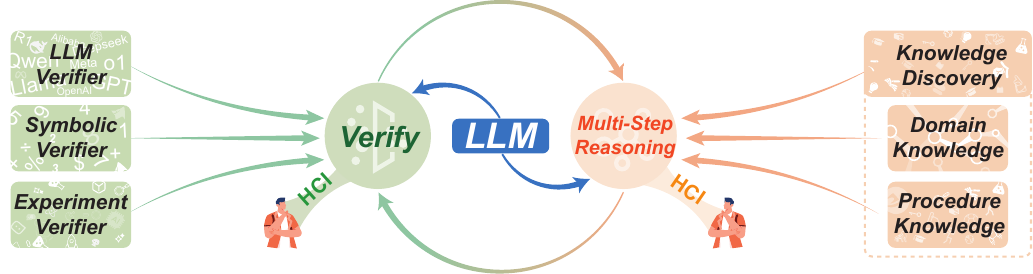}
\caption{The loops of complex problem solving.}
\label{fig:cps}
\end{figure}

Figure \ref{fig:cps} illustrates LLM-based techniques for complex problem solving. 
Current Chain-of-Thought (CoT) LLMs are trained through data synthesis. The process begins with generating CoT data, followed by selecting the correct CoT samples using a verifier for model training. During inference, the LLM generates multiple CoT solutions, and a verifier is used to identify the correct one for the given task.
There are multiple approaches to synthesizing data. One method involves having the LLM generate CoT data autonomously, which requires the base model to be well-trained. For applications with limited training data, knowledge mining can be conducted on existing datasets to synthesize data, while human expertise can also be incorporated. Additionally, the mined knowledge can be injected into the LLM during inference rather than being solely used for training.
Certain applications produce results that are difficult to verify, such as machine learning tasks. In such cases, multiple verification methods can be employed. Besides using an LLM-based verifier, symbolic verification and experimental evaluations can be conducted. 
Additionally, human experts may also be involved in the verification process.

\subsection{Multi-step reasoning} \label{sec:cot}

Chain-of-thought reasoning with LLM has been proven to be effective for solving complex problems.
% The initial paper of CoT
This line of research started from \cite{wei2022chain} that shows chain-of-thought
prompting with a few examples of reasoning paths can enhance the reasoning capabilities of LLM.
% Let's think step by step
\cite{kojima2022large} later demonstrates that chain-of-thought reasoning can improve
performance in a zero-shot setting by simply encouraging LLM to generate intermediate reasoning
steps with the ``Let's think step by step'' prompt.
% self-consistency
\cite{wang2023self} shows that sampling multiple reasoning paths and using majority vote
can further improve the performance of LLM in reasoning tasks.
% Tree of thoughts
\cite{yao2023tree} introduces Tree of Thoughts (ToT) that allows LLMs to explore
multiple reasoning paths over thoughts to improve the reasoning capabilities of LLMs.

We can utilize the architecture depicted in Figure \ref{fig:cps} to improve
chain-of-thought reasoning for tackling complex problems.
When a question is
posed, a \textit{generator} powered by a large language model (LLM) produces
multiple reasoning paths. These paths are then evaluated by a \textit{verifier}
to determine their accuracy. If some of the reasoning paths are validated as
correct, they are used to formulate an answer to the question. However, if none of
the paths are deemed correct, a \textit{corrector} is employed to create
new reasoning paths by modifying the incorrect ones and incorporating additional
feedback from the verifier. In this approach, enhancing the likelihood of getting
a correct solution for any given question requires improving two key metrics:
\begin{itemize}
    \item \textbf{coverage}: the percentage of problems that can be solved using
    at least one of the generated reasoning paths,
    \item \textbf{precision}: the probability of selecting the correct reasoning path from
    all generated paths.
\end{itemize}
\noindent To boost coverage, we need to fine-tune both the \textit{generator} and the \textit{corrector} to increase the chances of producing a valid reasoning path.
To enhance precision, the \textit{verifier} must be fine-tuned to more accurately identify the correct path.

% synthetic data generation to improve LLM for reasoning.
\paragraph{Generator.}
To refine the generator, we must move beyond human-produced data and instead synthesize
data with reasoning paths. \cite{zelikman2022star} present an iterative procedure that
generates multiple reasoning paths, selects the correct ones, and uses them to further
fine-tune the LLM, progressively improving its ability to produce accurate reasoning.
Additionally, they introduce a "rationalization" technique that leverages the problem's answer as a hint to enhance the generation of reasoning paths.
\cite{singh2024beyond} takes a similar iterative approach of generating reasoning paths
to fine-tune LLM and using fine-tuned LLM to generate more reasoning paths. The main difference
is that this work generates reasoning paths by using temporature sampling to
generate multiple paths and use a binary reward function to score them, while
\cite{zelikman2022star} uses greedy decoding to generate reasoning paths.
Both works show that the LLM model overfits quickly with the generated data.
\cite{bansal2024smaller} shows that it is not necessary to use a strong LLM to generate
high-quality synthetic data and the models fine-tuned on data generated by weaker LLMs
can consistently outperform those trained on data generated by stronger models.

% Methods for self-correction
\paragraph{Self-correction.}
We can use the incorrect reasoning paths from the previous attempt and the feedbacks from
the verifier to increase the probability of generating a correct one in the next iteration.
This process is considered as \textit{self-correction}.
% self-correction
\cite{huang2024large} shows that the existing LLM, such as GPT-4 and Llama-2, cannot increase
the probablity of generating right paths when being used for self-correction. They tend to
decrease the probability of getting right solutions compared with standard prompting methods.
This indicates that we need a specifically fine-tuned LLM for self-correction.
Pair-SFT \cite{welleck2023generating} trains an independent corrector model to
refine the outputs of a generator model.
They create a dataset consisting of response pairs ($y$, $y'$), where $y$ is
the initial response to a question and $y'$ is the corrected version, to train
the corrector.
SCoRe \cite{kumar2024training} employs a reinforcement learning approach to train a single LLM that both generates the initial response and self-corrects it. They find that previous methods are ineffective due to distribution shifts or the amplification of biases from the base model. By using a single LLM for both response generation and correction, SCoRe avoids the distribution mismatch that arises between separate generator and corrector models.

% How to search for a right reasoning path with inference compute resources
\paragraph{Inference scaling law.}
It is costly to generate many reasoning paths and select a correct one.
The more difficult a problem is, the more reasoning paths we may need
to sample.
A key research question here is how to use the compute resource
wisely to find a right path for any given problem.
% scaling law for inference-time compute
\cite{brown2024large} illustrates the scaling law for inference-time compute. They observe
that the coverage grows nearly log-linearly with the number of samples
generated from LLM and may reach to 100\% coverage if many reasoning paths are generated.
They further discover that it may be more cost-effective to generate
more samples with a weaker model than using larger LLMs when solving some simpler problems;
however, stronger LLMs are preferred when solving more difficult problems.
% how to scale LLM test-time compute optimally
\cite{snell2024scaling} investigates the ``compute-optimal'' strategy for scaling
inference-time compute in multiple aspects when generating a right reasoning path.
% different search strategy
When using a reward model to search for a good reasoning path,
they evaluated different search strategies, including best-of-N search, beam search and
lookahead search, and conclude that beam search is preferable for harder problems and lower computational budgets, while best-of-N is more effective for easier problems and higher budgets.
% sequential inference vs. parallel inference
Another aspect is to update the proposal distribution of the generator model to increase
the probability of generating a good reasoning path. An option is to generate multiple
reasoning path in parallel, while the other is to use a fine-tuned
LLM to iteratively revise their own answers, which results in a sequential test-time inference.
They show that easier questions benefit from sequential inference while harder
problems require some ratio of sequential to parallel inference.
% inference vs. pre-training
Compute resources can also be allocated to pre-training. To solve hard problems,
some compute resources should be used for pre-training while for easier problems,
we only need to use compute resources for inference.

\subsection{Knowledge} \label{sec:methods:knowledge}
Solving complex problems requires leveraging knowledge effectively. 
On one hand, many complex problem solving tasks are inherently domain-specific, and without specialized knowledge, solving them effectively becomes a challenge. 
On the other hand, the process of tackling these tasks involves multifaceted procedures, and existing large language models often lack the reliability and robustness required. 
Consequently, acquiring and enhancing this type of specialized knowledge represents a critical issue for effective complex problem solving.

To acquire this type of knowledge, the simplest and most direct approach is domain-specific pre-training \cite{song2025injecting}. 
While LLMs acquire world knowledge during training, research has shown that they are unreliable in memorizing and applying such knowledge—particularly long-tail information \cite{sun2024head}—to practical tasks, and multiple studies \cite{ren2024learning, gekhman2024fine} indicate that they are unable to acquire new factual knowledge through supervised fine-tuning (SFT) after pre-training.
Unlike these approaches, prompt-based enhancement techniques—such as RAG \cite{gao2023retrieval}, GraphRAG \cite{edge2024local,han2024retrieval}, and KAG \cite{liang2024kag}—can directly embed domain knowledge into the context of a specific task. 
Building on this, many studies have explored acquiring such knowledge through methods like information extraction \cite{zhang2019long,zhang2021document,xu2024large}, constructing domain-specific knowledge graphs \cite{zhang2021alicg,zhong2023comprehensive,ding2024automated}, or procedure extraction \cite{ye2025longproc}, as well as directly generating task-specific workflows with large language models \cite{qiao2024benchmarking,zhang2024aflow} and refining them through human interactive feedback \cite{boussioux2024crowdless}.
The following sections introduce various works categorized by the type of knowledge they address.

%Therefore, it is crucial to carefully manage knowledge utilization when handling tasks in new domains and enable LLMs to incorporate external knowledge for solving complex problems.

%\zd{we should include works on knowledge mining. Knowledge graph construction is one example of knowledge mining, but the current knowledge graph is not constructed for complex problem solving.}

\paragraph{Domain Knowledge.} 
Domain knowledge is designed to provide prior information for complex tasks, offering comprehensive introductions, detailed descriptions, and relevant background \cite{chen2022knowprompt,gu2024xiezhi,peng2024domain}.
\cite{lieto2019beyond} proposes a computational framework that enhances agents' problem-solving abilities by integrating a  goal-directed dynamic knowledge generation mechanism.
\cite{wang2023knowledge} introduces Knowledge-Driven Chain-of-Thought (KD-CoT), a framework that leverages external knowledge to verify and refine reasoning traces, thereby mitigating hallucinations and reducing error propagation.
\cite{lichain} introduces Chain-of-Knowledge (CoK), an innovative framework that enhances large language models by dynamically integrating grounding information from diverse sources.
\cite{pang2024physics} proposes Physics Reasoner, a knowledge-augmented framework that leverages large language models to solve physics problems.

\paragraph{Procedure Knowledge.} Procedure Knowledge refers to workflows or cognitive patterns designed to address complex problems, typically used to standardize and guide the reasoning processes of large models.
%Additionally, researchers are trying to models supplied a priori knowledge to bolster LLM's inferential capabilities.
Techniques like MoT~\citep{emnlp23_mot} leverage synthetic and extracted high-quality thought processes as external memory, providing the model with superior problem-solving examples.
Further, the BoT~\citep{agent_bufferofthought} paradigm introduces meta-buffers that store cross-task cognitive templates encompassing general reasoning patterns and knowledge structures, which can be reused and instantiated across various specific problems, thereby enabling efficient reasoning.
Furthermore, methods like Expel~\citep{aaai24_expel} also involve collecting an experience pool through environmental interactions, wherein the model learns from similar experiences and contrasts successful and unsuccessful trajectories to gain new insights and improve task inference capabilities. 
\cite{zhu2024knowagent} introduces KnowAgent, an approach that boosts the planning capabilities of LLMs by integrating explicit action knowledge.
Additional research by ~\cite{agent_workflow_mem, zhang2024aflow} utilize workflows that selectively guide agents for complex problem solving.

%wkm

\paragraph{Human-computer Interaction.}
Even with an external knowledge base, LLMs can still struggle with nuanced or domain-specific information, often lacking the deep contextual understanding that human experts possess. 
To address this, LLMs can collaborate with humans to bridge this gap by enabling humans to provide critical insights, ensuring that the LLM focuses on relevant information and refines its interpretations based on specialized knowledge.
For instance, in tasks like legal or medical research, humans can guide LLMs to prioritize certain references or nuances that the model might overlook~\citep{stiennon2020learning,amodei2016concrete,wu2021humanintheloopdeepreinforcementlearning,Huang_2024}. To enable such human-LLM collaboration, we need to design intuitive, user-friendly interfaces that facilitate effective communication and interaction between humans and LLMs~\citep{do2024facilitating,zhang2025breaking}. These interfaces should enable effective bidirectional communication, where users can provide feedback, clarify ambiguous inputs, and track the LLM’s reasoning in real-time. 
A well-designed interface fosters trust, enhances collaboration, and ensures that the LLM can be effectively used by both experts and non-experts alike.

\subsection{Evaluation} \label{sec:eval}
When tackling complex problems, it is essential to evaluate the effectiveness of solutions to enhance the reliability of LLM-based systems and identify better approaches. Prior research \citep{shi2023large, mirzadeh2024gsm} has demonstrated that LLMs are easily distracted by irrelevant information in mathematical reasoning. This suggests that LLMs may not truly grasp mathematical concepts but rather rely on pattern matching to generate responses. Additionally, \cite{mccoy2023embers} highlights that LLMs perform worse on rare tasks than on more frequent ones, even when the tasks share the same level of complexity. Moreover, LLMs are sensitive to the probability distribution of inputs and outputs in their training data (Internet text), even for deterministic tasks. These challenges become even more pronounced when LLMs are applied to domain-specific problems that are less commonly found online. To comprehensively assess solutions, multiple evaluation criteria—such as correctness and efficiency—may be considered. Ensuring that a solution meets practical requirements necessitates the development and integration of diverse evaluation techniques to effectively analyze LLM-generated solutions.

% Use a verifier to select the right reasoning path.
To improve chain-of-thought reasoning, we need a verifier model for selecting a right reasoning path.
This was first demonstrated by \cite{cobbe2021training} in solving math problems in GSM8K.
This work shows that training a verifier model to select a right solution among multiple solutions
can significantly improve the test solve rate compared with just fine-tuning LLM.
Therefore, a key problem here is how to train a reliable verifier model to determine the right
reasoning path.
% process supervision
\cite{lightman2024lets} shows the effectivenss of using process supervision to train
a process reward model (PRM). This method first generates multiple reasoning paths for
a problem and has human labelers to annotate labels for each individual step of the reasoning
paths. This approach requires many human resources to prepare for the training data.
\cite{lightman2024lets} adopts active learning to reduce human labeling efforts.
% process supervision without human labeling.
\cite{wang2024math} proposes a method that eliminates the need for human labeling
when training a PRM. 
To assess the correctness of intermediate steps in a reasoning path, they use a fine-tuned LLM to generate multiple subsequent reasoning paths from a given step. 
The correctness score for that step is then determined by the number of paths that successfully lead to the correct answer.
% generative verifier
\cite{zhang2024generative} trains a generative model as a verifier and shows that a generative model outperforms a discriminative verifier.
In addition, they show that training a single LLM for both generation and verification
can outperform separate LLMs for generation and verification.

% Tools as verifier
In addition to LLM-based verification, tools are employed to validate model outputs, mitigating hallucinations and enhancing accuracy. These verification methods can be broadly categorized into \textit{symbolic verification} and \textit{experimental verification}. 
         
Symbolic verification uses formal methods to ensure the correctness of the outputs of LLMs. This includes generating executable code and verifying its syntax and semantics through compilation~\citep{icml23_pal,tmlr23_pot}. Additionally, outputs are compared against knowledge bases or knowledge graphs to validate factual accuracy. These methods are especially effective for tasks that require logical consistency, such as mathematical proofs or domain-specific fact-checking~\cite{agent_factool}. PAL~\citep{icml23_pal} uses symbolic reasoning to interpret natural language problems and generate programs as intermediate steps. These programs are verified in runtime environments, like Python interpreters, ensuring that the logic and structure of the generated code are valid and executable.
In mathematical reasoning, tools like those in~\cite{nips23_mathtool} provide specialized interfaces for numerical computation, equation solving, and expression transformation. These interfaces allow the model to verify and correct each step, ensuring the correctness of the reasoning process, much like symbolic theorem proving. Factool~\cite{agent_factool} provides a flexible, domain-independent framework designed to identify factual errors. 
It enhances fact verification across domains by utilizing multiple verification tools, including search engines, academic databases, and code interpreters. 
         
In contrast, experimental verification involves validating models through real-world testing and empirical experiments \cite{arxiv24_mlebench,arxiv24_autokaggle,grosnit2024large}. This approach is useful when formal verification is impractical or when the goal is performance optimization. Models are tested in practical environments or simulations, with performance measured against benchmarks or competing solutions. 
In automated data science, frameworks like AutoKaggle \cite{arxiv24_autokaggle} exemplify experimental verification. These models autonomously participate in Kaggle competitions, optimizing data analysis pipelines and achieving top-tier performance by iterating through real-world testing, model tuning, and comparative analysis. \citet{grosnit2024large} orchestrates structured reasoning to automatically analyze and optimize solutions, while \citet{arxiv24_autokaggle} uses a multi-agent framework to generate, test, and refine models. 

For critical applications, ensuring safety and robustness is critical when applying LLMs in high-stakes or unpredictable environments, where incorrect outputs can lead to severe consequences. LLMs, while powerful, can generate unreliable or unsafe responses due to hallucinations, misinterpretations, or unexpected inputs. In this case, we should introduce human oversight to validate and correct outputs, ensuring safer and more reliable decision-making. For instance, in medical diagnosis, human experts can verify AI-generated treatment recommendations to avoid misdiagnoses or unsafe prescriptions~\citep{hakim2024needguardrailslargelanguage,zöller2024humanaicollectivesproduceaccurate}.

%% file: apps.tex
This paper examines four domains of real-world applications where LLMs can be applied to solve complex problems in these domains: software engineering, mathematics, data science, and scientific research. We will discuss the challenges in these applications from the perspective of multi-step reasoning, knowledge integration and result verifications.

% We discuss the domains as follows:
% the main tasks in these problems;
% the unique challenges in the three perspectives
% the current works and how these works address the challenges
% potentially future works

\subsection{Software Engineering}
This involves enabling LLMs to perform complex software engineering tasks with minimal human intervention. The core tasks in this domain are generally categorized into two major areas: code generation and code understanding. Code generation encompasses program synthesis \citep{jiang2024survey,li2024mmcode, zhang2024o1, tian2024codehalu, li2025structured}, code translation \citep{pan2023understanding,yan2023codetransocean,chen2018tree}, automatic program repair \citep{Jiang2023ImpactOC,Huang2023AnES,Ribeiro2023GPT3PoweredTE}, and code optimization \citep{du2024mercury,yan2023codescope}, where LLMs must produce functionally correct and efficient code that satisfies diverse specifications. Code understanding, on the other hand, focuses on analyzing and interpreting existing code, involving tasks such as source code summarization \citep{sym14030471,screenspotpro,wang2020cocogum}, code review \citep{yang2024survey}, and code search \citep{10.1145/3656341,du2021single}. Although these tasks differ in goals, they both demand that LLMs deeply understand the syntax, semantics, and structure of codebases and reason across multiple levels of abstraction.

Solving complex software engineering tasks using LLMs presents several unique challenges. First, these tasks require multi-step reasoning, as software development often involves decomposing problems, maintaining contextual consistency across files or functions, and iteratively refining code. Second, knowledge integration is essential--LLMs must possess foundational programming knowledge (e.g., syntax, algorithms), domain-specific practices (e.g., tool usage, design patterns) as well as the large code repository. Third, result verification is nontrivial: generating syntactically correct code is insufficient; it must also compile, execute correctly, and meet performance goals. Unlike natural language tasks, software correctness can be formally tested, creating both an opportunity and a challenge for using execution feedback effectively.

To address these challenges, a variety of models and frameworks have been proposed. For program synthesis, methods such as Code Evol-Instruct \citep{luo2023wizardcoder} and OSS-INSTRUCT \citep{wei2023magicoder} enhance LLM capabilities through synthetic data generation and fine-tuning, while approaches like GraphCoder \citep{liu2024graphcoder} and GALLa \citep{zhang2024galla} inject structural representations (e.g., code graphs) to improve syntactic and semantic understanding. Feedback-based mechanisms―such as Self-Debugging \citep{chen2023teaching}, LDB \citep{zhong2024ldb}, and RLTF \citep{liu2023rltf}--use runtime outputs, compiler errors, or test cases to iteratively guide model refinement. In repository comprehension, tools like StarCoder2 \citep{lozhkov2024starcoder}, DeepSeek-Coder \citep{guo2024deepseek}, SWE-GPT \citep{ma2024lingma}, and RepoCoder \citep{zhang2023repocoder}, CoCoMIC\citep{ding2022cocomic} and RepoFuse\citep{liang2024repofuse} utilize repository-level information, dependency graphs, and retrieval-augmented generation (RAG) to help models navigate large and interdependent codebases. For code optimization, frameworks like PIE-Problem \cite{ye2024iterative} and SBLLM \citep{gao2024search} introduce multi-programmer solution sets and evolutionary search strategies to help LLMs learn from diverse optimization techniques and refine code based on execution metrics.

Future work in automating software engineering will likely focus on three directions. First, building stronger reasoning-aware models that can generate and revise code through intermediate abstractions, such as pseudocode or symbolic plans. Second, enhancing long-context and memory mechanisms to handle complex repositories and cross-file reasoning. Third, incorporating closed-loop feedback systems that integrate automated test generation, runtime profiling, and formal verification into the code generation process. By combining these approaches, we can expect LLM-based agents to evolve from basic code assistants into capable autonomous software engineers.

\subsection{Mathematics}
Mathematical reasoning has emerged as a crucial benchmark for evaluating the capabilities of LLMs, as it requires not only natural language understanding but also precise logical reasoning, symbolic manipulation, and deep domain knowledge \cite{romera2024mathematical,sun2025challenging}. The main tasks in this field include arithmetic computation problems \citep{imani2023mathprompter,liu2023goat, yang2024arithmetic,guo2023arthmodel}, math word problems (MWPs) \citep{srivatsa2024challenging,heyue2023symbolicsolvers,gaur2023symbolic,kim2023longmwp}, and automated theorem proving (ATP) \citep{ahn2024llms,Yang2025Carts}.
These tasks test core competencies such as computational accuracy, deductive reasoning, the ability to model real-world scenarios mathematically, and the application of formal mathematical knowledge. Together, they serve as a rigorous framework for assessing whether LLMs can go beyond surface-level language generation to engage in structured, rule-based problem solving.

However, solving mathematical problems presents unique challenges that distinguish it from other complex domains. One major challenge is multi-step reasoning, as many mathematical tasks require sequential and logically dependent operations where one misstep can derail the entire solution. Another critical challenge is knowledge integration--LLMs must not only understand abstract principles (e.g., induction), but also domain-specific concepts and theorems, and recognize when and how to apply them, especially in the graduate level and research. This requires retrieving and manipulating domain-specific knowledge that is usually long-tail knowledge for LLMs. A third challenge is result verification, especially in settings like theorem proving, where the correctness of a result can only be confirmed through human evaluations or rigorous formal checking. Recent studies \cite{mahdavi2025brains} have shown that the current state-of-the-art LLMs generate correct final results but incorrect solutions in mathematical competitions. These challenges demand more than just fluent text generation--they require models to reason with precision, incorporate external tools or knowledge bases, and verify the correctness of multi-step solutions.

To address these challenges, recent research has introduced a range of specialized strategies and systems. For computational ability, models such as MathGLM \cite{yang2023gpt} are pre-trained on progressively complex mathematical problems using curriculum learning, achieving superior accuracy even compared to larger general-purpose models. Prompting-based methods like MathPrompter \cite{imani2023mathprompter} improve accuracy in arithmetic by generating and cross-verifying multiple solution paths. In reasoning tasks, symbolic integrations \cite{yang2024arithmetic} with Prolog or proof assistants like Lean (e.g., LeanDojo \cite{yang2024leandojo}, AlphaProof \citep{deepmind2024imo}) help bridge the gap between informal reasoning and formal logic to verify the mathematical reasoning generated by LLMs. In modeling and abstraction, efforts such as symbolic solvers for MWPs and autoformalization benchmarks (e.g., LeanEuclid) \cite{murphy2024autoformalizing} illustrate how LLMs can map real-world problems or geometric reasoning into formal mathematical representations. Moreover, retrieval-augmented systems and knowledge-grounded toolkits like DOCMATH-EVAL \cite{zhao2024docmath} and LeanDojo \cite{yang2024leandojo} show that integrating structured mathematical knowledge significantly boosts performance in tasks that require prior theorems or domain-specific reasoning strategies.

Looking forward, future work in LLM-based mathematical reasoning may focus on deepening the model’s ability to conduct formal reasoning with external feedback and process supervision. Developing hybrid frameworks that combine LLMs with theorem provers, symbolic execution engines, or even formal verification compilers could further enhance result correctness and logical soundness. Additionally, enriching LLMs with structured mathematical knowledge bases, improving their ability to retrieve relevant prior knowledge, and training them on fine-grained proof steps could enhance their capacity for advanced mathematical reasoning. Ultimately, achieving generalizable, verifiable, and domain-aware mathematical reasoning will be key to pushing LLMs closer to human-level mathematical understanding.

\subsection{Data Science}
This is a field where we perform data analysis and data modeling on a large amount of data \cite{zhang2024benchmarking}.
The main tasks in data science revolve around a complex, multi-stage pipeline that includes task understanding, data exploration and analysis, feature engineering, model selection, model training and evaluation. Each of these stages is interrelated, requiring not only technical execution but also careful reasoning and adaptation based on the input data. Unlike domains where problems are well-defined and static, data science demands continuous adjustments to explore the input data.

The unique challenges in this domain stem from its dynamic and data-dependent nature.
First, multi-step reasoning is essential, as decisions made in early stages (e.g., feature extraction) significantly affect later ones (e.g., model performance). Second, effective solutions often require domain-specific knowledge that is not easily captured by general-purpose LLMs; integrating such knowledge is vital to handle real-world complexity. Third, verifying the quality of a solution is particularly difficult because the performance depends heavily on the input data rather than just problem descriptions. This makes it challenging to assess modeling strategies.

Current research efforts have made significant progress in addressing these challenges through the development of agent-based systems. 
Data Interpreter \cite{hong2024datainterpreter} introduces a graph-based agent that models dependencies between pipeline stages and automates code generation and refinement accordingly. AutoKaggle \cite{arxiv24_autokaggle} employs a multi-agent framework―featuring specialized agents such as the Planner, Developer, and Reviewer―to provide end-to-end solutions for tabular data tasks, complete with iterative debugging and testing. Agent K \cite{grosnit2024large} optimizes performance through learned memory mechanisms, using reinforcement signals to retain useful strategies for future tasks. Meanwhile, DS-Agent \cite{guo2024dsagent} takes a knowledge-based approach by building a repository of expert insights derived from Kaggle competitions and applying case-based reasoning to generate better solutions. These systems are benchmarked using platforms like DS-Bench \cite{arxiv24_dsbench}, MLE-Bench \cite{arxiv24_mlebench}, and MLAgentBench \cite{mlagentbench}, which provide structured tasks rooted in real-world ML challenges to evaluate performance across the entire modeling pipeline.

Looking ahead, future research in this domain should focus on enhancing LLMs' ability to reason, adapt, and learn from data-driven experimentation. One key direction is the development of knowledge-enriched modeling agents that can incorporate advanced, domain-specific techniques beyond commonly used libraries. Another promising area is the integration of experiment-driven reasoning, enabling agents to iteratively test, evaluate, and refine their modeling strategies based on actual performance metrics. Finally, training LLMs with chain-of-thought (CoT) mechanisms that incorporate feedback loops from experiment results and domain-specific cues may offer a path toward more intelligent and adaptive data science agents.
%Such advances would bring us closer to creating general-purpose AI systems capable of tackling complex, open-ended modeling tasks with real-world impact.

\subsection{Scientific Research}
Artificial intelligence (AI) are increasingly playing a transformative role in scientific research, supporting tasks such as data analysis, simulation, literature review, and idea generation. Their application spans numerous domains, including biology, where tools like AlphaFold \cite{AlphaFold2021} and RoseTTAFold \cite{baek2021accurate} revolutionized protein structure prediction; physics, where AI assist in accelerating particle simulations \citep{kumar2023acceleratingparticlefluidsimulations}; and astronomy, where they aid in exoplanet detection \citep{miao2024ai}. In these contexts, LLMs are primarily used in scientific research in two ways: as tools that enhance human research capabilities and as co-creators that propose novel scientific hypotheses or ideas.

Despite these advances, the use of LLMs in scientific discovery presents several notable challenges. First, scientific research typically involves open-ended problems with unclear goals, making it difficult to apply LLMs in ways that guarantee accurate or verifiable solutions. Moreover, scientific research requires deep domain-specific knowledge, and LLMs must effectively leverage this expertise to make reliable predictions. These challenges make it difficult for LLMs to autonomously navigate the full research cycle, especially when the tasks involve open-ended reasoning, abstract synthesis, or interdisciplinary knowledge.

Due to the challenges of scientific research, LLMs are mainly used as tools to assist scientific tasks. For example, LLMs have been employed to accelerate data interpretation in fields like biomedicine and environmental science, where pre-trained models such as BioBERT and SciBERT help contextualize domain-specific data \citep{lee2020biobert,beltagy2019scibert,lo2019s2orc,jansen2023leveraging}. In simulation and predictive modeling, LLMs have been applied to climate forecasting and molecular modeling, leveraging their world knowledge to support scenarios where traditional simulations may be limited \citep{brown2020language}. For literature review and synthesis, LLMs help researchers uncover trends and identify knowledge gaps by summarizing extensive textual corpora \citep{scherbakov2024emergence,susnjak2024automating, beltagy2019scibert,jin2023pubmed,lo2019s2orc,luo2024large}. More experimental efforts use LLMs for research idea generation--some studies show that LLMs can generate novel scientific ideas, but also highlight the difficulties of evaluating and selecting high-quality ideas, especially as LLMs themselves are not reliable evaluators \cite{si2024llms, wang2024scipip, wang2024scimon, baek2024researchagent}. Additionally, agent-based systems like AI Scientist \cite{lu2024aiscientist} and HEADS \cite{su2024heads} demonstrate the feasibility of automating entire research pipelines, from idea generation to simulated peer review, though they fall short of validating these pipelines in solving truly difficult, real-world scientific problems.

Future research will likely focus on improving the reliability and impact of LLMs in scientific discovery by integrating more rigorous evaluation mechanisms and enabling deeper domain-specific reasoning. One key direction is building multi-agent collaboration frameworks that mimic scientific team dynamics to diversify and refine generated ideas. Another is combining LLMs with external tools--such as experiment databases, simulation engines, or formal verification systems--to support result verification and reduce hallucinations. Finally, improving the feedback loop between LLM-generated outputs and human or experimental validation will be critical for realizing LLMs as trusted collaborators in the scientific process. These developments will help move from speculative generation toward verifiable, impactful contributions to scientific research.

%% file: discussions.tex
Although there has been notable progress in LLM research for solving complex problems, significant challenges persist. To further enhance LLMs' ability to tackle complex problems, we should focus on improving them from the three key perspectives: multi-step reasoning, knowledge and verification.

\paragraph{Multi-step reasoning.}
There are two major problems in training LLMs for multi-step reasoning: \textit{data scarcity}
and \textit{high computation cost}.

CoT LLMs are typically pre-trained on vast amounts of Internet data, with further enhancement achieved through synthetic data generated by LLMs. However, data scarcity is still a challenge in many specialized domains. For instance, while widely-used programming languages like Python have large code corpora available online, lesser-known languages such as Lean \cite{demoura2015lean} suffer from limited data.
Although synthetic data generation via LLMs can improve LLMs, it relies on the base LLMs being well pre-trained for the specific domain. 
As a result, using data synthesis to improve an LLM’s ability to generate code in languages like Lean remains a significant challenge.
Similar issues arise in other fields, including mathematics and science.
One approach to tackle the data scarcity problem is to develop agents that combine LLMs with custom models specifically trained for the target applications. 
For example, in formal theorem proving, where data is limited, custom models can help determine the applicability of mathematical strategies (tactics) and assess whether the proof has progressed towards the goal after each step \cite{Yang2025Carts}. 
These models guide LLMs in making informed decisions with reinforcement learning \cite{wang2025ragen,qian2025toolrl,wang2025otc}, enhancing their reasoning capabilities even in domains with sparse data.

Another problem is \textit{high computation cost}.
The inference scaling law has been identified as a way to enhance LLMs' ability to tackle complex problems \cite{chen2024not,sui2025stop,wang2025harnessing,feng2025efficient}. 
By generating numerous reasoning paths, LLMs are more likely to find a path that leads to a solution for highly complex problems at the cost of increased computation. 
For example, GPT-o1 and its successor GPT-o3 exhibit significantly higher inference costs compared to GPT-4. 
Therefore, it is crucial to reduce inference costs.
We can address the computational challenges from several angles.
First, we can train better LLMs for generation and self-correction to reduce the number of trials to generate reasoning paths/tokens \cite{slow}.
Second, we should explore various search algorithms to generate reasoning paths more efficiently. 
Besides best-of-N, we should also explore beam search and Monte Carlo Tree Search.
Third, we can reduce the model size of LLMs to speed up inference, which includes techniques like distilling LLMs into smaller models and decoupling knowledge from LLMs to create more compact versions, thus reducing the computational requirements.

%  随时切换fast  slow    multiagents    collective intelligence 

\paragraph{Knowledge.}
Knowledge is fundamental to solving complex problems. 
Currently, LLMs acquire world knowledge through next-token prediction on massive data during pre-training, which leads to several challenges.
LLMs may not reliably memorize world knowledge, particularly long-tail knowledge \cite{sun2024head}. 
As a result, the current LLM cannot work well on domains where training data are scarce.
Even if LLMs retain knowledge, they may struggle to recall relevant information when solving complex problems or LLMs may lack the ability to correctly apply knowledge to solve complex problems.

To effectively leverage knowledge in solving complex problems, one approach is to construct comprehensive knowledge graphs that go beyond traditional triplet-based structures, which contain only entities and relations.
In the context of machine learning, a specialized knowledge graph should include not only verbal descriptions of techniques but also their mathematical formulations and corresponding implementation code.
Additionally, it should capture relationships between different techniques to facilitate the exploration of diverse approaches and foster innovation in problem-solving.
Such a knowledge graph can be systematically built by extracting information from academic papers, technical reports, and textbooks, with careful validation and verification \cite{luo2024oneke}.
Once established, this knowledge graph can be utilized in two key ways. 
First, it can be used to synthesize data for model training, addressing data scarcity challenges. 
Second, it can support problem-solving during inference through a retrieval-augmented generation (RAG) approach, enabling the model to access and apply relevant knowledge in real time \cite{guo2024lightrag}.

However, large language models still face challenges in representing and discovering knowledge. 
Their reliance on chain-of-thought reasoning for complex tasks is hindered by current serialized techniques, which struggle to structurally capture domain-specific knowledge and logic (workflow) while providing limited support for human intervention \cite{min2025self}. 
Additionally, large language models encounter difficulties in balancing innovative knowledge discovery with logical credibility, often leading to hallucinatory outputs. 
Compounding these issues, the large language model' dynamic adaptation capabilities are insufficient to keep pace with rapidly changing environments, as delayed knowledge updates  can render decision-making strategies ineffective.
These interconnected challenges underscore the need for further research into improving thought process modeling, enhancing domain knowledge discovery, updating (editing) \cite{yao2023editing,wang2024knowledge,zhang2024comprehensive,du2025rethinking}, and developing more robust adaptation mechanisms for complex problem solving \cite{fu2025agentrefine,xue2025illusion}.

\paragraph{Evaluation.}
The current LLM research, such as OpenAI o1, focuses on complex problems whose final results can be easily verified, such as competitive coding and mathematical reasoning. However, practical applications have much more complex requirements that complicate the verification of final results.
First, some applications not only require correctness of solutions but also to achieve efficiency or yield better accuracy. In machine learning tasks, for example, while baseline methods like random forecasts or multilayer perceptrons can be considered as "correct" solutions, they may not meet the desired performance, and more effective solutions are preferred.
Furthermore, problems in many applications are difficult to define comprehensively. Again using machine learning tasks as an example. Both the task description and the distribution of input data are essential for designing an effective solution. However, conveying the data distribution of the input data to an LLM is challenging.
In addition, in certain scientific fields, such as drug discovery, climate modeling, or social sciences, validation of results often requires extensive experimental testing, replication, or further theoretical analysis to confirm their accuracy and reliability.

These challenges emphasize the need for robust evaluation frameworks and the integration of domain-specific expertise to ensure the reliability of LLM-generated outputs. To enhance the credibility of LLM outputs, it is crucial to employ multiple evaluation approaches.
Consider machine learning tasks as an example—there are several ways to assess the effectiveness of a machine learning algorithm:
% based on previous experiments.
First, the algorithm’s performance can be evaluated by comparing it with previously published results, such as academic papers and technical reports.
% based on LLM's evaluation.
Secondly, an LLM-based evaluator can be utilized to assess the quality of a solution. To improve its accuracy, data analysis should be conducted to extract comprehensive insights from the input data and feed them to LLM.
% based on experiment evaluation.
Thirdly, implementing the machine learning algorithm and conducting experiments provides an empirical assessment of its effectiveness.
% based on theoretical analysis
Fourth, for certain machine learning algorithms, we can perform some theoretical analysis on the algorithm, which is further verified by symbolic verification tools like Lean, ensuring rigorous validation of the algorithm’s correctness and effectiveness.
By combining all these different evaluation approaches, we can potentially perform thorough evaluation of machine learning algorithms.
We believe similar evaluation principles (the combinations of LLM-based evaluations, empirical experiments, theoretical evaluations) can also be applied to other domains.

%% file: related.tex
Several survey papers have addressed LLM-based reasoning. Early works, such as \citet{qiao2023reasoning} and \citet{huang2023towards}, provide an overview of LLM-based reasoning, which is crucial for complex problem-solving. However, these surveys primarily focus on the initial developments in this area. With the release of GPT-o1 \cite{openai2024gpto1}, the effectiveness of LLM-based reasoning was significantly demonstrated. Following this, numerous studies have explored the potential mechanisms behind GPT-o1. For example, \citet{zeng2024scaling} and \citet{xu2025large} delve into techniques that could enable o1-like reasoning, particularly through reinforcement learning.
In contrast, this work takes a broader perspective, addressing all the different capabilities needed for complex problem-solving, rather than focusing solely on reasoning.

Numerous survey papers focus on specific domains of LLM-based reasoning. For instance, \citet{yang2024formal} examines the progress, challenges, and future directions in formal mathematical reasoning. \citet{eger2025transforming} explores recent advances in using LLMs to support scientific research, covering applications such as literature search, idea generation, text and multimodal content (e.g., scientific figures and diagrams) generation, and AI-based peer review.
\citet{ahn2024large} provides an overview of various types of mathematical reasoning using LLMs. However, this work does not address o1-like techniques. Meanwhile, \citet{li2024survey} focuses on theorem proving within mathematical reasoning. Instead of solely relying on LLMs, this survey breaks down theorem proving into multiple components and discusses the application of various deep learning techniques for each aspect.

%\zd{we need to include some knowledge graph LLM surveys}

%% file: conclusion.tex
In this survey paper, we define complex problem-solving from the perspectives of both cognitive science and computational theory, and analyze the characteristics of different complex problems. We then investigate significant advancements in large language models (LLMs), with a focus on chain-of-thought reasoning and agent-based approaches in the context of complex problem-solving. We discuss how data synthesis and reinforcement learning have enhanced LLMs' abilities for multi-step reasoning. Additionally, we explore how the agent-based approach enables AI systems to harness external knowledge, tools for execution and result verification. However, we also examine the limitations of these methods when applied to different types of complex problems.

%We further explore how LLMs can enhance complex problem-solving across four key domains: software engineering, data science, mathematics, and scientific research. While each domain stands to gain significantly from the capabilities of LLMs, they also present unique challenges. For example, in software engineering, LLMs are widely recognized for code generation. To better solve software engineering tasks, they must also have the capability of navigating large codebases to understand code; in addition, they not only need to generate correct code, but also more efficient code as well. In data science, LLMs can automate tasks like data cleaning, feature engineering, and model selection, but they currently tend to generate code based on the most commonly used algorithms. Generating machine learning algorithms that are specifically suited to the problem at hand remains a challenge. In mathematics, LLMs can assist in the derivation and proof processes, particularly in handling complex formulae and theorem proofs. However, they still face challenges in understanding abstract concepts, ensuring accuracy, and performing creative reasoning. Finally, in scientific research, LLMs are used as a tool to assist scientists in tackling open-ended problems, facilitating human-LLM collaboration in the discovery process. Each of these domains requires tailored strategies to fully leverage the potential of LLMs while addressing the associated challenges.